\documentclass[journal,twocolumn]{IEEEtran}
\usepackage{amsmath, graphicx, color, booktabs, multirow}

\newcommand{\revised}[1]{{\color{black} #1}}

\begin{document}

\title{Markerless Body Motion Capturing for 3D Character Animation based on Multi-view Cameras}
\author{Jinbao~Wang, Ke~Lu,~\IEEEmembership{Member,~IEEE,} and~Jian~Xue~\IEEEmembership{Member,~IEEE}%
    \thanks{Manuscript received XXX XX, 2019; revised XXX XX, 2019. This work was supported in part by the National Key R\&D Program of China under contract No. 2017YFB1002203, the National Natural Science Foundation of China (NSFC, Grant No. 61671426, 61731022, 61871258,  61572077), the Beijing Natural Science Foundation (Grant No. 4182071), the University of Chinese Academy of Sciences (Grant No. Y95401YXX2), the Instrument Developing Project of the Chinese Academy of Sciences (Grant No. YZ201670), and the Scientific Research Program of Beijing Municipal Education Commission (Grant No. KZ201911417048). (\emph{Corresponding author: Jian Xue.})}%
    \thanks{J.B. Wang, K. Lu, and J. Xue are with University of Chinese Academy of Sciences, Beijing 100049, China (e-mail: linkingring@163.com, luk@ucas.ac.cn, and xuejian@ucas.ac.cn).}} 

\IEEEtitleabstractindextext{
\begin{abstract}
This paper proposes a novel application system for the generation of three-dimensional (3D) character animation driven by markerless human body motion capturing. \revised{The entire pipeline of the system consists of five stages: 1) the capturing of motion data using multiple cameras, 2) detection of the two-dimensional (2D) human body joints, 3) estimation of the 3D joints, 4) calculation of bone transformation matrices, and 5) generation of character animation.} The main objective of this study is to generate a 3D skeleton and animation for 3D characters using multi-view images captured by ordinary cameras. The computational complexity of the 3D skeleton reconstruction based on 3D vision has been reduced as needed to achieve frame-by-frame motion capturing. The experimental results reveal that our system can effectively and efficiently capture human actions and use them to animate 3D cartoon characters in real-time.
\end{abstract}
	
}

\maketitle
\IEEEdisplaynontitleabstractindextext
\IEEEpeerreviewmaketitle

\section{Introduction}
\IEEEPARstart{T}{he} study of human body motion capturing has made considerable progress over recent decades, and has been driven by the practical requirements of entertainment, sports, and clinical applications. 

According to a classification scheme presented in \cite{Moeslund2006A}, it is possible to separate those applications into three groups according to their main objective: surveillance, control, and analysis.
First, control applications interpret motion from an actor and transform it into a sequence of operations, such as video game animation \cite{Magnor2015Digital}.
Second, analysis applications generally study a person as a set of objects for applications in sports, biomechanics, and rehabilitation \cite{Bezodis2008Lower, Churchill2015The, Hiley2012Achieving, Neil2007Contributions}.
Third, motion capturing is a powerful technique for animating virtual characters and controlling them for purposes such as character animation \cite{Baran2007Automatic}.

Most previous studies have required high quality cameras or a chromatic background to precisely segment the person in the foreground. Various recent methods have used depth sensors to improve the efficiency of data acquisition. However, these methods have focused on the body shape or skeleton reconstruction, which are typically expensive and unsuitable for ordinary applications.

\begin{figure*}[htb]	
	\centering
	\includegraphics[width=0.98\linewidth]{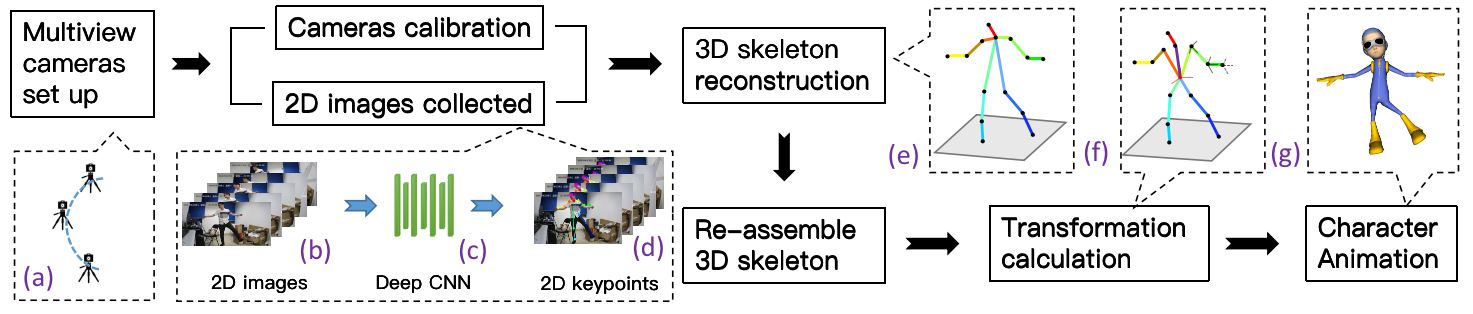}
	\caption{
		The processing pipeline. 
		(a) multi-view cameras;
		(b) images captured;
		(c) 2D joint keypoints detector; 
		(d) output of 2D joint keypoints; 
		(e) 3D skeletion;
		(f) reassembled 3D skeleton and bone transformation; 
		(g) animation of a 3D character.
	}\label{fig:pipeline}
\end{figure*}

To satisfy the requirement of various applications, such as 3D animation and virtual reality (VR) games, for simple and easy-to-use motion capturing techniques, this paper proposes a markerless body motion capturing system based on multi-view cameras. This system can use a few inexpensive cameras to generate a body skeleton and effectively animate the characters in five stages, \revised{as shown in Fig. \ref{fig:pipeline}}.

The first stage is the capturing of motion data using multiple cameras, and includes the cameras’ calibration and synchronization.
After the capturing of multi-view images, the CNN pose detector detects the 2D skeleton in the second stage.
The third stage is the calculation of the 3D skeleton based on the multi-view geometry constraint.
In the forth stage, the bone transformation is calculated using the 3D skeleton in preparation for the final stage.
Finally, character rigging is achieved using the 3D skeleton obtained in the final stage and subsequently optimized to vividly animate the 3D character.

\revised{The main contributions of this paper can be summarized as follows. First, a practical method is proposed to obtain the 2D keypoints of the human body in multi-view images from ordinary cameras based on deep CNN. Second, a new accurate and robust 3D joints estimation method is proposed to overcome the problems caused by cameras’ errors or incorrect 2D joints detected by CNN. In addition, a method for calculating bone transformation matrices is also presented to generate a full 3D skeleton frame-by-frame automatically without registering and avoid character’s unnatural behavior in the animation, e.g. bone spinning. Finally, a practical system is designed and implemented for the generation of 3D character animation through markerless human body motion capturing by cheap consumer-grade devices, e.g. the ordinary network cameras.}

The rest part of this paper is organized as follows:
related work is introduced in Section \ref{sec:related}, including some commercial and academic approaches for motion capturing. 
Our approach is described overall in Section \ref{sec:approach}. 
Designed experiments and their results with analysis are presented in Section \ref{sec:experiments}. 
Finally, conclusion and future work are drawn in Section \ref{sec:conclusions}.

\section{Related Work}\label{sec:related}

Motion capturing is the process of recording the movement of objects or people. In recent years, many systems and approaches have been proposed for capturing human activities, and can be categorized in two types, as follows. 

\emph{Marker-based motion capture.}
Currently, marker-based motion capturing is a mature technique that has been used successfully in many fields such as the motion picture industry and VR. 
However, this method is disadvantaged by the controller having to be worn as a marker suit with sensors
such as optical markers \cite{Raskar2007Prakash} or mounted cameras \cite{Shiratori2011Motion}. Thus, this makes it impossible to capture the movement of people wearing ordinary clothes.
Additionally, marker based motion capturing is sensitive to skin movement relative to the underlying bone \cite{Della2005Human}.
Moreover, the exact placement of markers on anatomical landmarks is difficult to realize, and markers placed on the skin do not directly correspond to the 3D joint positions. Presently, many commercial automatic optoelectronic systems, developed by companies such as The Captury, Organic Motion, and Simi, can be used to investigate human motion. 
However, these system mostly employ multi-view cameras that can deduce the 3D position of the objects and body skeleton.
Although the specifications of these systems have noticeable differences \cite{Richards1999The}, 
the same underlying principles apply in terms of several points of interest being located in sequential images, converted to real-space coordinates, and used to infer the 3D pose of the underlying skeleton.

\emph{Markerless motion capture.}
A markerless system that can address the limitations and eliminate the need of body-worn sensors has attracted a substantial amount of attention in the field of computer vision and computer graphics, and has expanded the application of human motion capturing.
Recently, this field has increasingly attracted more interest from researchers.
The four major components of a markerless motion capturing system are (1) the camera systems being used, (2) the representation of the human body (the body model), (3) the image features being used, and (4) the algorithms used to determine the parameters (shape, pose, and location) of the body model \cite{Colyer2018A}.
There are two types of camera systems for markerless motion capturing, and can be distinguished according to whether or not a “depth map” is estimated. 
Currently, motion capturing systems based on depth-sensing are considered as an effective method of estimating a full-body pose in real-time, for example, in \cite{Shotton2011Real, Ye2013A}.

From an algorithmic viewpoint, markerless motion reconstruction can be classified into two main categories; discriminative approaches \cite{Ganapathi2010Real, Shotton2013Real, Cao2016Realtime} and generative approaches \cite{Gall2009Motion, Bregler2004Twist, Duetscher2000Articulated}.

\emph{Discriminative approaches.}
In this field, the idea behind discriminative algorithms is to convert the motion capturing problem into a regression or pose classification problem, so as to achieve a much faster processing time, improve robustness, and reduce the dependence on initial guesses. 
Discriminative algorithms can be divided in two major groups. 
One approach is to discover a mapping directly from the image features to a description of the pose, such as by using machine-learning-based regression \cite{Agarwal2005Recovering, Hong2016Multi}.
Alternatively, a dataset of pose examples can be created and then searched to discover the most similar known pose if the current image is given, as has been done in previous studies \cite{Chen20113D, Babagholami2014A, Theobalt2017VNect}.
However, discriminative algorithms have reduced accuracy and require a very large set of training data, which makes their application difficult.
Therefore, the discriminative approaches are typically used as the initial guesses in generative approaches \cite{Rhodin2016General}.

\emph{Generative approaches.}~
In the generative motion capturing approaches, the pose and shape of the body are acquired by fitting the model to information extracted from images.
This can generate a set of model parameters such as the body shape, bone length, and joint angle.
Alternatively, the 3D body model can be compared against a 3D reconstruction, such as a visual hull, by minimizing the distances between the 3D vertices of the model and the 3D points of the visual hull \cite{Corazza2010Markerless, Corazza2006A} through a standard algorithm known as the iterative closest point.
Contrary to discriminative algorithms, generative approaches are typically based on temporal information and can solve a tracking problem. 
The majority of these methods parameterize the high dimensional human body and embed a low dimensional skeleton into the body model.
Then, the motion reconstruction process can be formulated as a frame-by-frame optimization to deform the skeletal pose \cite{Stoll2011Fast} and surface geometry \cite{De2008Performance, Guo2015Robust}, alone or simultaneously \cite{Vlasic2008Articulated, Hasler2009Markerless, Ye2012Performance, Liu2011Markerless, Liu2013Markerless}.

Baran \sl{et al.}\em~\cite{Baran2007Automatic} has presented a method of automatic character animation, wherein the skeleton is applied to the character and attached to the surface, which then allows the skeletal motion data to animate the character. 
Film \sl{et al.}\em~\cite{Flam2009OpenMoCap} proposed an open source system for optical motion capturing called OpenMoCop, which has been developed based on digital image analysis techniques.
Many methods can be used to obtain the human body joints, such as human 3D scanning \cite{Liu2013Markerless, Guo2015Robust, Xu2016FlyCap}, reconstruction \cite{Liu2013Markerless, Guo2015Robust, Xu2016FlyCap} or CNN-based methods \cite{Theobalt2017VNect, Pavlakos2017Harvesting}, for example. However, a complete system for animating virtual avatars or characters is currently lacking.
Additionally, obtaining a fine representation of the entire body through 3D reconstruction is difficult and computationally expensive. Hence, such a system is very hard to implement in commercial applications. Moreover, it is challenging to achieve satisfactory performance when using the CNN-based method to acquire the 3D joints.

\revised{This study focuses on the 3D reconstruction of the body's skeleton  the animation of a virtual character.} A complete practical system is proposed which can easily and rapidly capture the motion of the human body without using markers.

\section{Approach}\label{sec:approach}

This section describes our technical approach in detail. 
Our objective is to realize markerless body motion capturing that is better suited to ordinary participants compared with inconvenient marker-based capturing equipment, such as Mocap.

The pipeline of our system mainly includes five stages, as shown in Fig. \ref{fig:pipeline}. 

\begin{itemize}
	\item Stage 1 (a-b): \revised{multiple ordinary network-cameras} shown in (a) are used to collect the 2D multi-view images shown in (b), and are jointly calibrated.  
	\item Stage 2 (b-d): (c) is the 2D human body joint detector that detects the joint keypoints in the 2D images using a deep convolutional neural network (Deep CNN); (d) produces the multi-view output images with the 2D joints.
	\item Stage 3 (d-e): the joint keypoints are calculated in 3D space according to the multi-view geometry constraint.
	\item Stage 4 (e-f): character rigging is achieved by reassembling the new 3D skeleton with the bone transformation matrix in (f) according to the skeleton in (e), which only contains the joint keypoints.
	\item Stage 5 (f-g): the bone transformation obtained for rigging is used to animate the 3D character.
\end{itemize}

\subsection{Capturing of 2D images from calibrated multiview cameras} \label{sec:stage_1}

First, the experimental environment is set up for body motion capturing. 
After their positions are fixed, all cameras are calibrated using fundamental matrix estimation for pairs of images. This is followed by bundle adjustment, which is a feature-based 3D reconstruction algorithm.
By applying bundle adjustment \cite{Triggs1999Bundle}, we achieve the reconstruction of the 3D sample space, and obtain all of the camera parameters, including the intrinsic matrix and extrinsic matrix.  
Additionally, the above mentioned calibration procedure only has to be carried out once for each camera layout.

\begin{figure}[th]
	\centering
	\includegraphics[width=0.99\columnwidth]{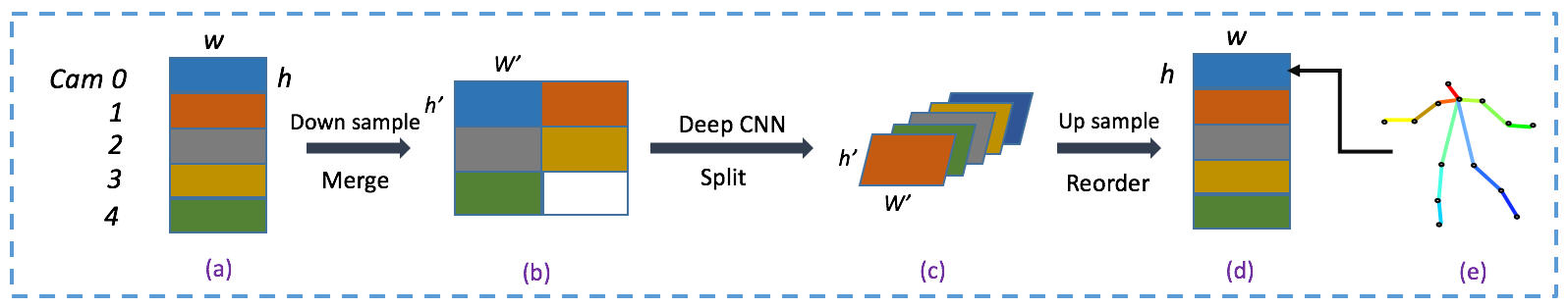}
	\caption{
		2D joint keypoint detection. 
		(a) input 2D images from multi-view cameras;
		(b) merge multiple 2D images into a single image; 
		(c) obtain out-of-order 2D joint keypoints from the Deep CNN detector;
		(d) reorder the 2D joint keypoints according to the camera id;
		(e) obtain the 2D skeleton for each camera image.
	}
	\label{fig:stage_a_d}
\end{figure}

\subsection{2D Joint detection from multi-view 2D images} \label{sec:stage_2}
Here, we describe our method of capturing the 2D human pose without landmarks, \revised{as shown in Fig. \ref{fig:stage_a_d}}.

In this \revised{part}, a deep CNN detector from a state-of-the-art open-source 2D pose estimation library \cite{Cao2016Realtime} is used to perform the detection of the 2D human pose keypoints.
As is well known, as the number of cameras increases, the obtained reconstruction accuracy becomes higher. Additionally, using the Deep CNN method is time- and resource-consuming.
Hence, using many cameras requires a substantial amount of computational resources.
Therefore, to balance the calculation speed and quality of 3D joint reconstruction, a procedure of merging multi-view images into a single image to be fed into the deep CNN is designed to speed up the calculation and improve the synchronization of the output 2D poses.

The details of the process are shown in Fig. \ref{fig:stage_a_d}. 
First, we down-sample all images and merge them into a single image. Next, by performing 2D pose detection, a series of 2D joints are obtained. Finally, the operation of \revised{\emph{upsampling}} and \emph{reordering} consist of scaling the 2D joint into the original input image size and matching the corresponding camera id.

After the above mentioned phases have been completed, we obtain the key joints of the 2D skeleton from all views.

\subsection{3D joint estimation using multi-view 2D joints} \label{sec:stage_3}
 
\begin{figure}[ht]
\centering
\includegraphics[width=0.45\textwidth]{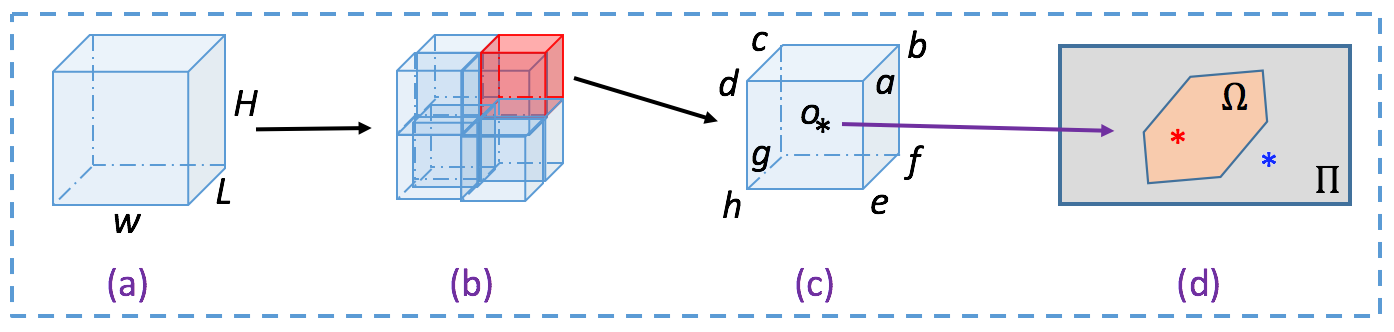}
\caption{
		Estimation of 3D joints using 2D joints.
		(a) a 3D sample space is initialized with width $W$, height $H$, and length $L$;
		(b) the space is subdivided;
		(c) one subspace is considered as a 3D sample unit;
		(d) the area $\Omega$ of the 3D sample unit is projected onto image $\Pi$.
}
\label{fig:stage_d_e}
\end{figure}

An overview of the 3D joint estimation method is shown in Fig. \ref{fig:stage_d_e}. The basic idea behind our approach toward estimating the 3D joints is based on the feature point correspondence and visual hull method. Compared with other methods for the dense reconstruction of complex dynamic scenes from multiple wide-baseline camera views, in this method, we simply reconstruct the sparse 3D points based on the matched 2D joint keypoints from the pose detection network. In our method, the space is subdivided to speed up the search for joint candidates in 3D space, as shown in Fig. \ref{fig:stage_d_e}. This approach consists of the following \revised{steps}.

\subsubsection{3D sample point projection onto 2D image area}

After starting the initial 3D space with $W$, $H$, $L$, the space is split into eight-fold 3D sub-regions, as shown in \revised{Fig. \ref{fig:stage_d_e}(a) and Fig. \ref{fig:stage_d_e}(b)}. Considering the 3D sample point shown in Fig. \ref{fig:stage_d_e}(c) as an example, nine points (eight vertices and one center point) of one sample cube, $\{a, \cdots\!, g, o\}$, are projected to the image $\Pi$ captured by the $i$-th camera. 
After projection by $\{a, \cdots\!, g\}$, the new points comprise the $\Omega$ area shown in Fig. \ref{fig:stage_d_e}(d).

\subsubsection{3D joint candidates and iterative space subdivision}

In the view of the $i$-th camera, a part of the area of the 2D image $\Pi_{i}$ is defined as $\Omega_{i}$, and a 2D point is denoted $p^{2}_{i}$.
Additionally, we defined a 3D sample cube with width $w$, height $h$, and length $l$ as $Cube_{\{w, h, l\}}$, and its center point with $p^{3}$.
$\operatorname{project}(p^{3}, \kappa)$ represents the projection from $p^{3}$ to $p^{2}$, where $\kappa$ denotes the camera parameters.
Thus, the projection \revised{area $ \Omega_i $} of the 3D sample cube in the $i$-th image is defined as follows:
$$\Omega_i=\{p^2\ \big\vert\ p^2 = \operatorname{project}(p^{3}, \kappa),\ p^3\in Cube\}.$$
Because the cube projection on the 2D plane is convex, its projection area can be calculated easily by its eight vertices \revised{$\{a, \cdots\!, g\}$}.

\begin{figure}[th]
	\centering
	\includegraphics[width=0.2\textwidth]{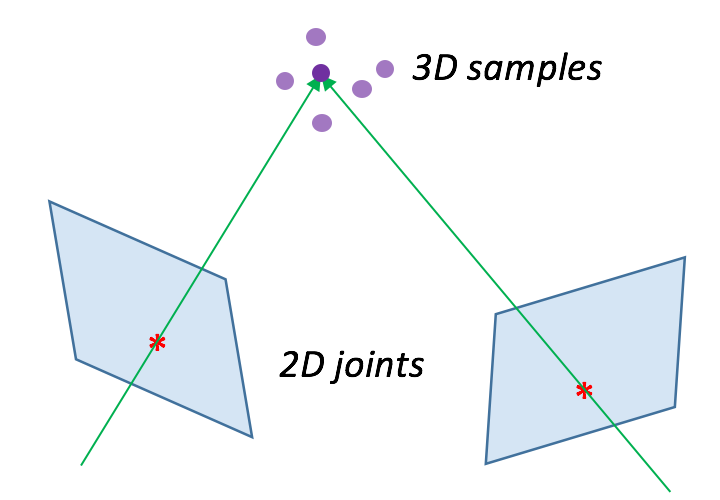}
	\caption{ 
		Projection that is a bounding geometry of the actual 3D point.
		Purple points are the 3D samples.
		Red stars are the 2D joints detected by the deep CNN.
	}
	\label{fig:projection}
\end{figure}

Accordingly, the foreground mask, known as a 2D joint keypoint, is the 2D projection of the corresponding 3D foreground object.
Along with the cameras’ viewing parameters, the 2D point $p^{2}$ defines a back-projected generalized line containing the actual 3D joint $p^{3}$, as shown in Fig. \ref{fig:projection}.

The projection is the bounding geometry of the actual 3D point.
The purple points indicate the 3D samples.
The red stars are the 2D joints detected by the deep CNN.
$\psi_{i}$ is defined as follows:
\[
\psi_{i} =  
\begin{cases}
  1 & \text{if } p^{2}_{i} \in \Omega_{i},\\
  0 & \text{otherwise}\\ 
\end{cases}  
\]
where we consider two cases for the relationship between the real 3D joint keypoint $p^{3}$ and its candidates.
If the $p^{2}$ detected in the 2D camera images is located in the projection area $\Omega$ of the $p^{3}$ candidates, the cube may contain its corresponding point $p^{3}$. Here, we define $N_{Cube} = \sum_{i=0}^n \psi_{i}$ under the multi-views meeting of  $\psi_{i}$.
Theoretically, information from at least two different perspectives is required to determine a point in 3D space.

An iterative space subdivision strategy is used to locate the keypoint of each joint in 3D space. This iterative process will continue until the stop condition occurs; that is, until $N_{Cube}<\sigma$ or size of $Cube < \delta$ is satisfied. In the stop condition, $\sigma$ is an integer constant, and is typically less than the number of cameras to ensure better robustness against error; that is, at least $\sigma$ camera images must satisfy the condition of $p^2_i \in \Omega_i$. Moreover, $\delta$ is the smallest cube size for space subdivision. 
The size of the cube is determined by the camera parameters, the 2D image resolution and the wide projection in the 2D image, which depends on the error of the pose detector.
In each iteration, the edge length of the cube for space subdivision is reduced by half, \revised{i.e. $Cube_{\{w, h, l\}} \rightarrow Cube_{\{w/2, h/2, l/2\}} $}.

\subsubsection{Clustering of 3D candidates} \label{clustering}

Because we set two threshold \revised{parameters} $\sigma$ and $\delta$, the collected 3D candidates are mainly distributed within a small range.
Therefore, we consider the average of the 3D candidates as the final 3D joints.
After clustering, we obtain the entire 3D skeleton as shown in Fig. \ref{fig:pipeline}(e).

According to the procedure described above, all joints of the captured person are processed one by one and their 3D candidates are finally obtained. The efficiency of the processing is ensured owing to the adoption of subdivision and sparse 3D keypoint reconstruction. 

\subsection{Transformation calculation of a character’s bone} \label{sec:stage_4}

Because it is difficult to provide the absolute transformation matrix for each joint in world coordinates, the transformation matrix is instead calculated in terms of the T pose in our method, which is called \emph{T pose template}.
Here, we introduce the method of calculating the bone’s transformation by using the 3D joints.
First, the body joints and bones are pre-defined.
Then, the 3D joint is used to compute the transformation of the bones.
Additionally, we propose a method for avoiding the self-spinning of the bone, which leads to an awkward and unnatural effect in the final animation.

\subsubsection{Definition of joints and bones}

\begin{figure}[th]
	\centering
	\includegraphics[width=0.4\textwidth]{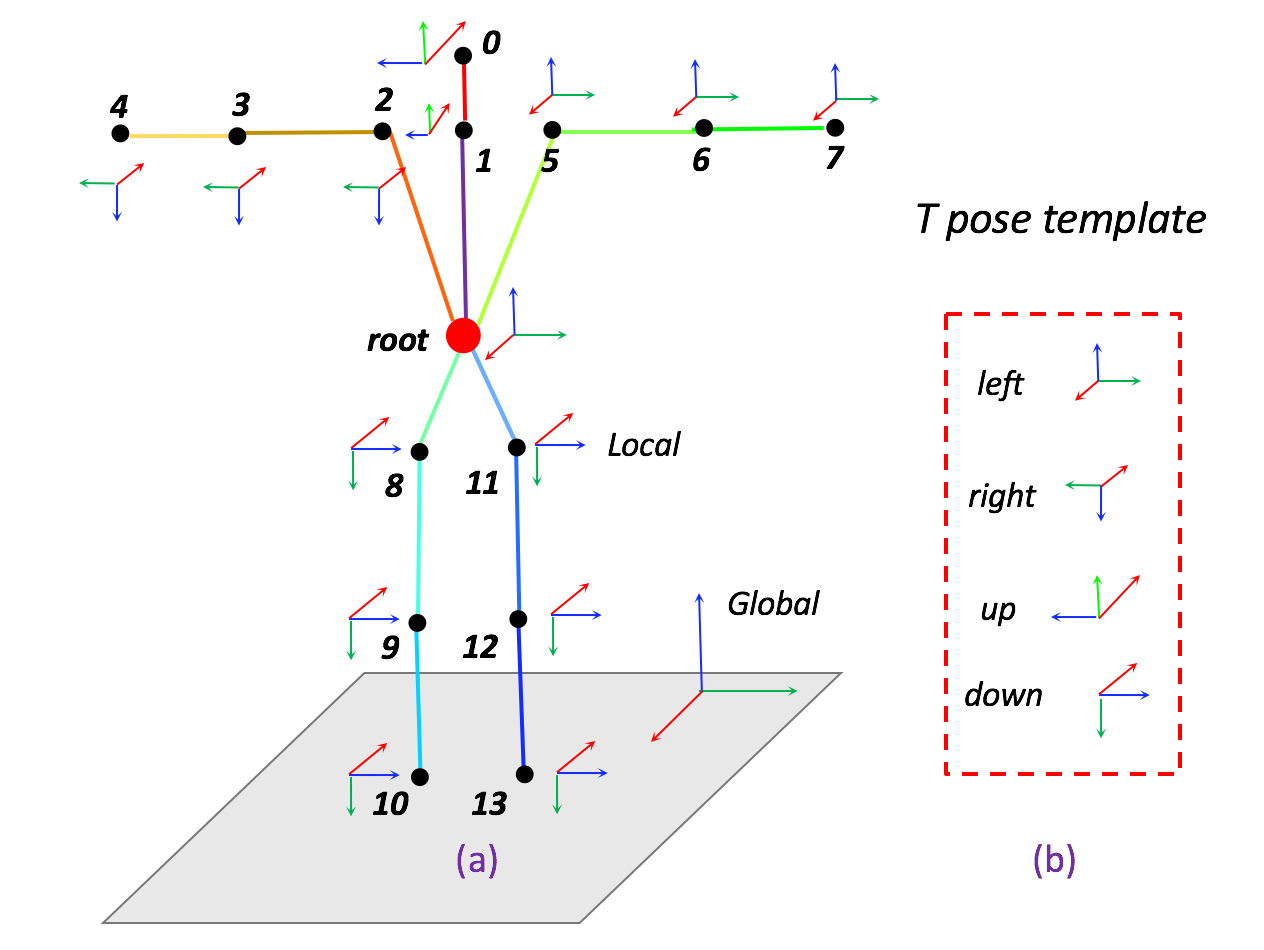}
	\caption{\label{T pose template}
		Template description of T pose. 
		(a) definition of the body's T pose template;
		(b) definition of local coordinates, the \emph{left}, \emph{right}, \emph{up}, and \emph{down}.
		} \label{fig:stage_e_f}
\end{figure}

The joints and bones of a 3D skeleton are defined in Table \ref{tab:bones definition} \revised{according to Fig. \ref{fig:stage_e_f}}.
We define the main joints of the human body; that is, the head, elbow, and knee, which can approximately express the body pose, whereas the subtle joints, such as the face or finger joints, are not considered.  
Additionally, we define a root joint called Torso to control the global movement and transformation.
The bone definitions are presented in Table \ref{tab:bones definition}. 
For example, the \emph{r\_upper\_arm} represents the bone of the right arm, which consists of the R\_Shoulder and R\_Elbow joints. 

Because our objective is to drive the movement of a 3D character, it is sufficient to use the location of the 3D joints only. 
Moreover, we must calculate the transformation of each bone to realize the movement of the 3D character.

Hence, we define the \emph{T pose template} coordinates before the transformation calculation, which includes the \emph{Global} and \emph{Local} coordinates shown in Fig. \ref{fig:stage_e_f}, wherein \emph{T pose template} defines the \emph{left}, \emph{right}, \emph{up}, and \emph{down} local coordinates.

\begin{table}[th]
	\setlength{\abovecaptionskip}{1mm}
	\centering
	\setlength{\tabcolsep}{2mm}
	\caption{\revised{A 3D skeleton joints and bones definition}}\label{tab:bones definition}
	\begin{tabular}{llll}
		\toprule
		Index    & Joint Name  & Indices & Bone Name            \\
		\midrule
		0        & Head        & 1, 0    & \emph{head}          \\
		1        & Neck        &         &                      \\
		2        & R\_Shoulder & 1, 2    & \emph{r\_shoulder}   \\
		5        & L\_Shoulder & 1, 5    & \emph{l\_shoulder}   \\
		3        & R\_Elbow    & 2, 3    & \emph{r\_upper\_arm} \\
		6        & L\_Elbow    & 5, 6    & \emph{l\_upper\_arm} \\
		4        & R\_Hand     & 3, 4    & \emph{r\_lower\_arm} \\
		7        & L\_Hand     & 6, 7    & \emph{l\_lower\_arm} \\
		8        & R\_Hip      &         &                      \\
		11       & L\_Hip      &         &                      \\
		9        & R\_Knee     & 8, 9    & \emph{r\_upper\_leg} \\
		12       & L\_Knee     & 11, 12  & \emph{l\_upper\_leg} \\
		10       & R\_Foot     & 9, 10   & \emph{r\_lower\_leg} \\
		13       & L\_Foot     & 12, 13  & \emph{l\_lower\_leg} \\
		root     & Torso       & root, 1 & \emph{torso}         \\
		\bottomrule
	\end{tabular}
\end{table}

Let $ x $ axis of the local coordinates point to the extending direction of the body part.
In \revised{Fig. \ref{T pose template}},
(a) is the template definition of the body’s T pose.
The \emph{left}, \emph{right}, \emph{up}, and \emph{down} are local coordinates, and are defined in the same manner as those in (b).
The \emph{root} coordinate is the same as the \emph{Global} coordinate.

\subsubsection{Transformation calculation}

The transformation matrix $T$ is defined as 
$$
	T_{4 \times 4} = \left[ \begin{matrix}
							R_{3 \times 3} & t_{3 \times 1} \\
							0_{1 \times 3} & 1 \\
							\end{matrix}
							\right].
$$
\revised{In matrix $T$}, $R$ is the $3 \times 3$ rotation matrix and $t$ is the $3 \times 1$ translation matrix.
Here, $t$ is obtained from the rigging information of the original character model. Additionally, the bone rotation matrix is calculated from the variation between the 3D joints, which were obtained as described in Section \ref{sec:stage_3}, and the T pose template.
To control the movement of the 3D character, the character model must already be rigged and skinned.
In the rigging and skinning phase, each bone transformation is initialized with a $R$ rotation matrix and $t$ translation matrix.
The $R$ rotation matrix controls the bone orientation, while the $t$ translation matrix controls the bone location in 3D space. 
The character’s bones with a pre-specified length are connected at the joints, and the rotation of the bones around these joints allows the positioning of the skeleton.
Therefore, there is no need to control the $t$ translation matrix using our own 3D joint data.
Here, $t$ is set by $[0, 0, 0]^T$.
The next task is to calculate the rotation matrix.

\begin{figure}[th]
	\centering
	\includegraphics[width=0.99\linewidth]{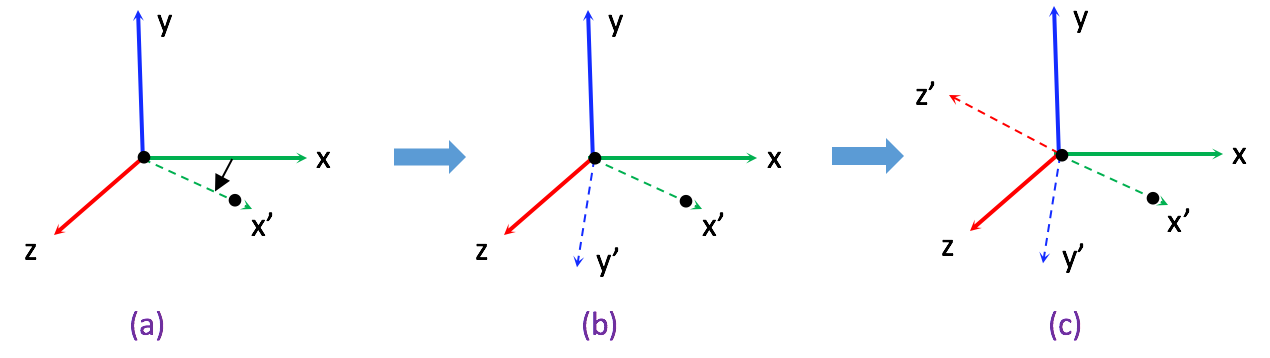}
	\caption{Illustration of the rotation matrix calculation; black points indicate neighboring joints: (a) $x'$ is the bone vector consisting of two points; (b) $y'$ is the cross product of $x'$ and $x$; (c) $z'$ is the cross product of $y'$ and $x'$.}\label{fig:rotation}
\end{figure}

\begin{figure}[th]
	\centering
	\includegraphics[width=0.4\textwidth]{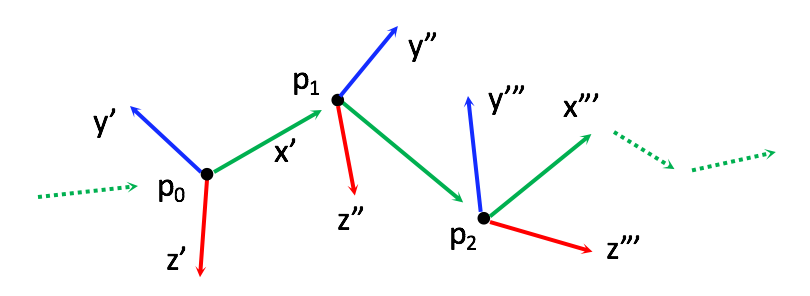}
	\caption{Bone and sub-bone calculation using the chain rule: $x^i$, $y^i$, $z^i$ are the coordinate axes; the black point $P_i$ indicates the joint.}\label{fig:chain}
\end{figure}

Fig. \ref{fig:rotation} illustrates the step of calculating the rotation matrix for one bone.
To estimate the rotation, we use the predefined standard coordinates. 
Therefore, one bone rotation can be calculated as follows.
$$
	\left\{ \begin{gathered}
	x' = x' \hfill\\
	y' = x' \times x \\ 
	z' = y' \times x' \\
	\end{gathered}  \right.
$$
$$R = [x'^T, y'^T, z'^T]. $$
When a sub-bone rotation $R'_n$ must be calculated after the bone rotation $R'_{n-1}$, we estimate it using the chain rule. 
In other words, the sub-bone rotation is calculated in terms of the bone coordinates.
Thus, the last joint has free rotation and no constraints.
The calculation step is shown in Fig. \ref{fig:chain}.

Fig. \ref{fig:chain} shows the chain rules for computing the sub-bone rotation matrix. 
The transformation for each bone is calculated by multiplying its bone rotation matrix with the parent bone's rotation matrix.
Therefore, the sub-bone and bone rotation $ R'_n $ are calculated as follows.
$$ R'_n = R_0 * R_1 * R_2 *....*R_{n}.$$

\begin{figure}[th]
	\centering
	\includegraphics[width=0.75\linewidth]{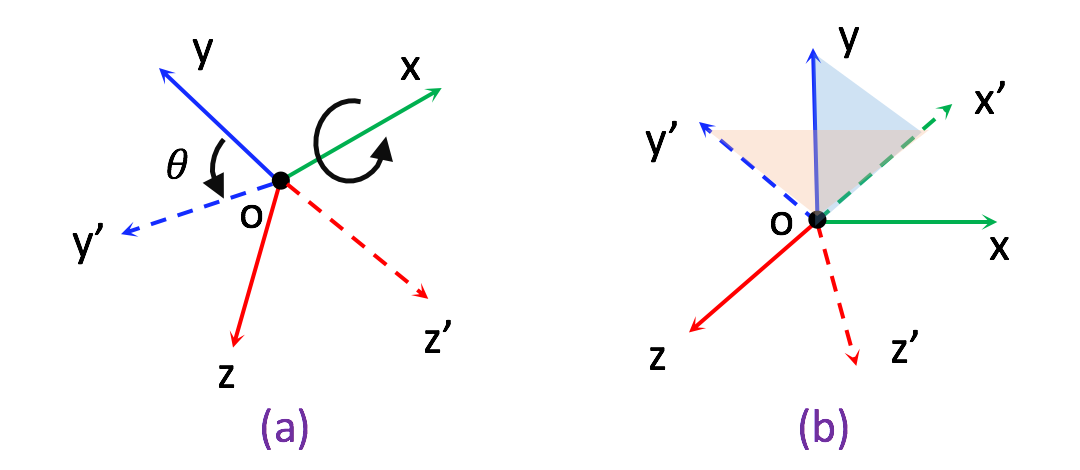}
	\caption{
			Illustration of coordinate spinning.
			(a) spinning of $x$-axis with rotation angle $\theta$;
			(b) coordinate transformation in calculation of spinning angle.
	}\label{fig:spinning}
\end{figure}

It is known that the method of using two 3D points for estimating the bone rotation matrix lacks constraints and leads to coordinate spinning around the $x$-axis, as shown in Fig. \ref{fig:spinning}(a).
Therefore, we propose a method of calculating the spinning angle to rotate the coordinate system backward around the $x$-axis.

The sub-bone coordinate can be expressed in the bone coordinate.
We assume that, when the bone rotates, the coordinates transfer by the $\Pi_{xoy}$ plane.
Then, we can calculate the spinning angle $\theta$, which is defined as follows:
$$ \theta = \angle(\Pi_{x'oy}, \Pi_{x'oy'}). $$
The $\angle(\cdot)$ is the angle of the two planes.
To avoid more cases wherein only one $\theta$ is appropriate for overcoming the spinning, an optimization-based method is used to satisfy the stop condition, wherein $y'$ is located on the $\Pi_{x'oy'}$ plane.
After obtaining the angle $\theta$, we use the Rodrigues' formula to rotate the coordinates and obtain the new rotation $R''_n$.

To ensure that the bone rotation has the same deformation effect in the global coordinates, 
we must use the coordinate transformation as follows to rewrite the final rotation matrix as $r_n$.
$$ r_n = R_{C} * R''_n * R_{C}^T, C \in \{left, right, up, down\}. $$
In the above expression, $R_{C}^T$ is the transpose of $R_{C}$.
The final bone transformation is expressed as follows:
$$
	T_{4 \times 4} = \left[ 
					\begin{matrix}
					r_{n} & t \\
					0 & 1 \\
					\end{matrix}
					\right].
$$

\subsection{Controlling the movement of a character} \label{sec:stage_5}

\begin{figure}[th]
	\centering
	\includegraphics[width=0.99\linewidth]{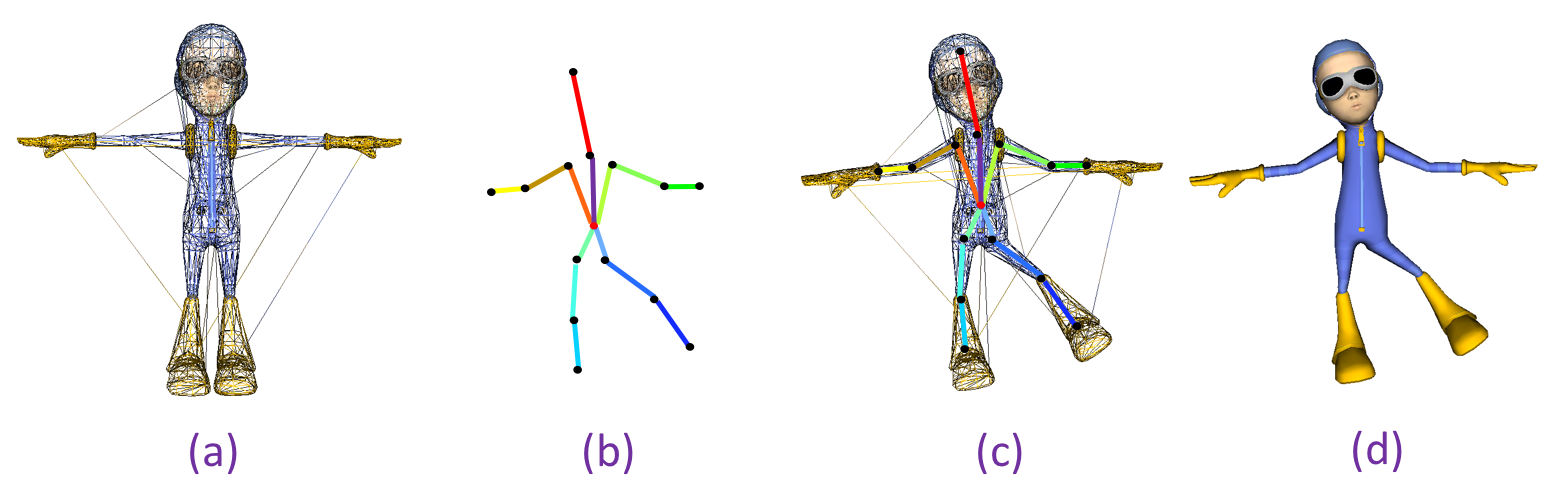}
	\caption{
			The character movement and process of binding the skeleton to the skin are as follows:
			(a) the character model is provided \sl{Astroboy}\em;
			(b) the skeleton is setup in the binding pose phase;
			(c) use the skeleton to obtain the shape (skinning);
			(d) the model is rendered using textures and materials.
	}\label{fig:driving}
\end{figure}

At this point, the skeleton can be used to transform the vertices of the modeling pose so as to position the character accordingly.
Here, we describe how bone transformation can be used to control the movement of a character.
The process is shown in Fig. \ref{fig:driving}.

The model named \sl{Astroboy}\em~shown in Fig. \ref{fig:driving}(a) has already been processed by modeling and binding.
Additionally, this could be used to draw the skeleton in the binding pose phase, as shown in in Fig. \ref{fig:driving}(b), and in the pose wherein the skeleton is attached to the skin, which is called skinning, as shown in Fig. \ref{fig:driving}(c).
Finally, the complete deformation of the character is obtained by rendering the model using textures and materials, as shown in Fig. \ref{fig:driving}(d).

\section{Experiments} \label{sec:experiments}

In this section, \revised{we present some experimental results with analysis of our method. Furthermore, the comparison with commercial optical motion capture system, such as ASUS Xtion PRO LIVE (Xtion), and other related methods is also included}. Our system is built with a workstation and five same network cameras (HIKVISION DS-IPC-B12-I). The workstation is equipped with a Intel Core i7-6850K@3.6GHz CPU, 32GB RAM and a NVIDIA TITAN X GPU.

\subsection{Comparison with Xtion} \label{sec:comparison with xtion}
 
\begin{figure*}[ht]
	\centering
	\includegraphics[width=0.95\textwidth]{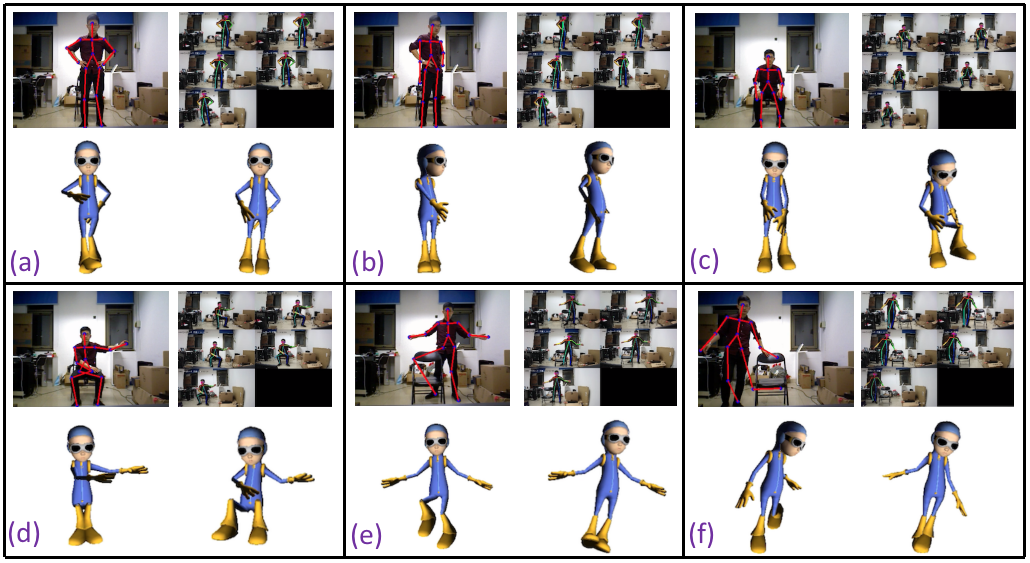}
	\caption{
			Xtion failure cases for which correct results were obtained by our system:
			(a) wrong bone transformation for right upper arm;
			(b) tracking failure (right upper arm);
			(c) tracking failure (with interference from other parts);
			(d) tracking failure (limb overlap);
			(e) tracking failure (objects in front of limbs);
			(f) tracking failure (objects very close to limbs);
	}\label{fig:xtion fail cases}
\end{figure*}

Here, we compare the performance of our system with the same functionality in ASUS Xtion PRO LIVE (Xtion), which carries a structure sensor and uses a 3D sensing solution provided by the PrimeSense company. PrimeSense developed the NiTE middleware, which analyzes the data from hardware and modules, such that OpenNI can provide gesture and skeleton tracking.

Compared with Xtion, our solution can obtain more accurate and robust results, as suggested by the examples shown in Fig. \ref{fig:xtion fail cases}. 
Because our system adopts a single-frame processing based method, whereas Xtion uses a tracking-based method, a slight jitter exists in the results produced by our system. Moreover, in some cases, Xtion fails to completely represent the skeleton’s joints in correspondence with the person captured by the camera, as shown in Fig. \ref{fig:xtion fail cases}. There are two main reasons for these failures: \revised{the tracking error and incorrect transformation matrix}. When the body limbs move quickly or are influenced by other objects, the motion tracking by Xtion will always fail, which produces a tracking error. Moreover, even if tracking is successful, Xtion may provide incorrect transformation matrices for some joints, which causes the animated character to behave strangely, as shown in Fig. \ref{fig:xtion fail cases}(a).

By using the non-tracking strategy and multi-view configuration, our system can reduce the interference of limbs or other objects and capture the body motion whenever possible. Moreover, matrix errors rarely occur in our system, owing to the precise method of calculating the 3D joints. Therefore, our system produces correct results for most cases wherein Xtion fails, as shown in Fig. \ref{fig:xtion fail cases}.

\subsection{Experiments on HumanEva-I dataset} \label{sec:humanevaexp}

We tested our system using the HumanEva-I dataset as a benchmark, which includes
several video sequences of people performing common actions, such as boxing, walking, and gesturing.
Additionally, the dataset was compiled with images captured using seven calibrated cameras (4 grayscale and 3 color) synchronized with the 3D body poses obtained from the motion capturing system.
\begin{figure}[htb]
	\centering
	\includegraphics[width=0.99\linewidth]{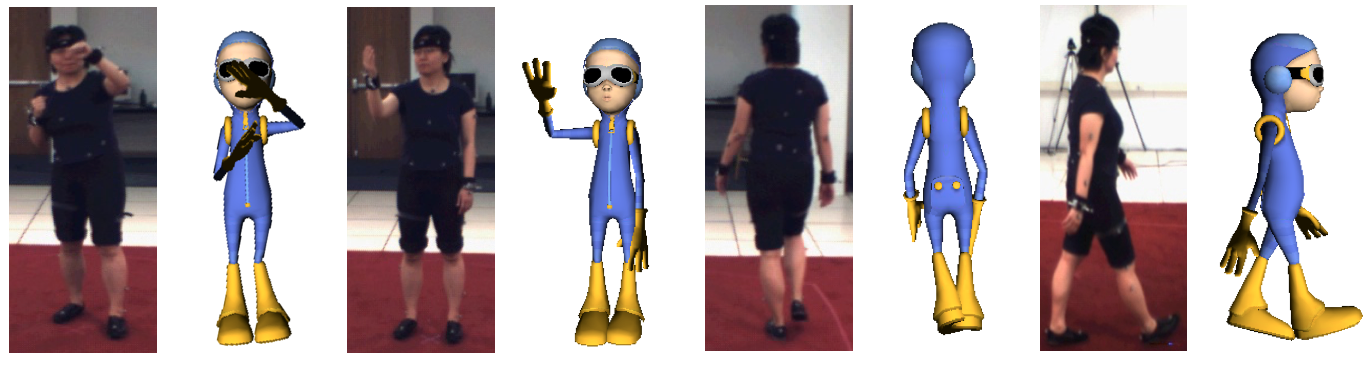}
	\caption{
		 Some experimental results on Humaneva-I dataset.
	}
	\label{fig:Humaneva examples}
\end{figure}
Fig. \ref{fig:Humaneva examples} shows examples of four actions in the HumanEva-I dataset, and the corresponding experimental results obtained by our system. 

Considering the body skeleton difference between the HumanEva-I dataset and our own data, we selected the joints corresponding to the location of the reflective markers.
We considered the 3D joints captured by mocap in the HumanEva-I dataset as the ground truth. Additionally, the error in the subsequent frames could be estimated as the distance from these points to their ground-truth positions. 
Therefore, the error metric employed was the average absolute distance between the real positions of the markers being tracked and the estimated positions.
Let $ x $ represent the pose of the body. 
We used the 3D error measure \cite{Cano2014Parallelization} in a single frame of the sequence, as follows:
\begin{equation}\label{formula_3d1}
	d^3(x, \hat{x}) = \frac{1}{n}\sum_{i=1}^n||x_i - \hat{x}_i||.
\end{equation}
In (\ref{formula_3d1}), $ n $ is the number of 3D joints, $ i = 1 \hdots n$, $ x $ is the 3D joint obtained by our system, and $ \hat{x} $ is the 3D ground-truth joint.
This equation was used to calculate the tracking error of a sequence as the average of all its frames.
For the sequence of $T$ frames we computed the average performance by using the following equation:
\begin{equation}\label{formula_3d2}
	\mu_{seq.} = \frac{1}{T}\sum_{t=1}^Td^3(x_t, \hat{x_t}).
\end{equation}
The mean absolute 3D error (\emph{Mean Abs 3D Err}) is computed using (\ref{formula_3d2}) in millimeters (mm).
In the experiment, we considered these joints (\revised{with indices 0-13 in Table \ref{tab:bones definition}}) to evaluate our method. The sampling cube size was set as presented in Table \ref{tab:action analysis}, which lists the \emph{Mean Abs 3D Err} in the sequences \emph{Gesture} and \emph{ThrowCatch} of \emph{S1} in the HumanEva-I dataset.

\begin{table}[th]
	\setlength{\abovecaptionskip}{1mm}
	\centering
	\setlength{\tabcolsep}{0.8mm}
	\caption{The \emph{Mean Abs 3D Err} reults of \emph{Gesture} and \emph{ThrowCatch} sequences of S1 in the HumanEva-I dataset}\label{tab:action analysis}
	\begin{tabular}{ccccccccc}
		\toprule
		 & \multicolumn{4}{c}{\emph{Gesture}}  & \multicolumn{4}{c}{\emph{ThrowCatch}}	   \\ 
		\cmidrule(r){2-5} \cmidrule(r){6-9}
		 $ \delta $ (mm)   & $ \sigma=3 $  & $ \sigma=4 $  & $ \sigma=5 $  & $ \sigma=6 $& $ \sigma=3 $  & $ \sigma=4 $  & $ \sigma=5 $  & $ \sigma=6 $      \\
		\midrule
		  $[5, 5, 5]$    & 50.7  & 49.5   & 41.7   & 44.2   & 48.0   & 50.8   & 49.2  & 53.3  \\
		  $[10, 10, 10]$ & 50.6  & 46.5   & 41.7   & 38.8   & 47.5   & 51.4   & 47.8  & 57.0  \\
		  $[15, 15, 15]$ & 50.6  & 46.5   & 41.6   & 38.8   & 47.5   & 51.4   & 47.8  & 57.0  \\
		  $[20, 20, 20]$ & 51.0  & 46.5   & 41.6   & 37.8   & 47.2   & 49.6   & 46.4  & 47.6  \\
		  $[25, 25, 25]$ & 51.0  & 44.2   & 41.6   & 37.8   & 47.2   & 49.6   & 46.4  & 47.6  \\
		  $[30, 30, 30]$ & 51.0  & 44.2   & 44.2   & 37.8   & 47.2   & 49.6   & 46.4  & 47.6  \\
	    \midrule
	      \emph{mean}  & 50.8  & 46.2   & 42.0   & 39.2   & 47.4   & 50.4   & 47.3  & 51.6  \\
		\bottomrule
	\end{tabular}
\end{table}

The \emph{Gesture} case shows that the \emph{Mean Abs 3D Err} value decreased when the view $ \sigma $ increased. Moreover, $ \delta $ was fixed, owing to the constraint from the multi-view geometry presented in Table \ref{tab:action analysis}. 
For example, when $ \sigma $ was fixed to 3 and $ \delta $ increased, the \emph{Mean Abs 3D Err} value  slightly increased because it reached the stop condition fast, and was thus stopped by the large sample cube size  $ \delta $.
However, when $ \sigma $ was fixed to 6, the \emph{Mean Abs 3D Err} value decreased as $ \delta $ increased.

This behavior can be explained as follows: when more views are needed to satisfy the geometry, the iterative stop condition is better formed. Therefore, a bigger coarse sample cube size $ \delta $ leads to more accurate results, as is the case for \emph{ThrowCatch}. 
Additionally, it is known that a small $ \delta $ \revised{leads to more time consumption}. 
Hence, we set $ \sigma=4 $ and $ \delta=[10, 10, 10] $ to achieve the most optimal trade-off with regard to accuracy and computing time.

The \emph{Mean Abs 3D Err} (mm) of the \emph{Gesture} and \emph{ThrowCatch} sequences in S1
for the results obtained with the HumanEva-I dataset are presented in Table III.
When the action was more intense and many interactive actions existed, the \emph{Mean Abs 3D Err} was higher, as in the case of the \emph{Jog} in S3 and \emph{Box} in S2, for example.

\begin{table}[th]
	\setlength{\abovecaptionskip}{1mm}
	\centering
	\setlength{\tabcolsep}{1.2mm}
	\caption{The \emph{Mean Abs 3D Err} results of sequences in the HumanEva-I dataset.}\label{tab:3d joints 3DErr}
	\begin{tabular}{cccccccc}
		\toprule
    Subject & $ \sigma $  & $ \delta $ (mm)   & \emph{Gesture}  & \emph{Box}  & \emph{Jog} & \emph{Walking} & \emph{ThrowCatch}  \\
		\midrule
		S1   & 4      & $[10, 10, 10]$ & 46.5  & 52.3   & 65.4    & 53.0    & 51.4        \\
		S2   & 4      & $[10, 10, 10]$ & 62.6  & 68.9   & 67.3    & 59.8    & 60.4        \\
		S3   & 4      & $[10, 10, 10]$ & 61.5  & 63.0   & 70.3    & 60.3    & -           \\
		\bottomrule
	\end{tabular}
\end{table}

\begin{figure}[htb]
	\centering
	\includegraphics[width=0.98\linewidth]{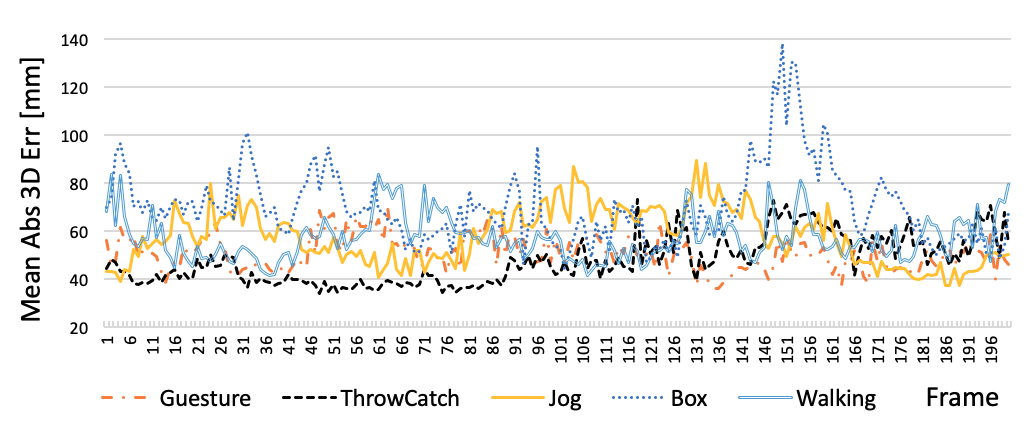}
	\caption{
		The \emph{Mean Abs 3D Err} results of sequences by frame in subject S1.
	}\label{fig:S1 sequences 3DErr}
\end{figure}

Fig. \ref{fig:S1 sequences 3DErr} shows the \emph{Mean Abs 3D Err} value of five sequences in subject S1.
In order to show the detail, we display 200 frames for each sequence.

Considering the result from a quantitative viewpoint, the errors were reasonably higher for many reasons: 
(1) The CNN detector introduced error and the obtained 2D joint positions were not very accurate. 
(2) The distortion of the cameras was not considered and this led to the position appearing incorrectly in the 2D images. 
(3) The ground truth of the HumanEva-I dataset was affected by large errors, for reasons such as the joint center not being precise, the markers not being located in the underlying skin, and the Mocap data not being very synchronous.
(4) There existed the error of joint positions between the ground truth and our definition, such as the head part.

\begin{table}[th]
	\setlength{\abovecaptionskip}{1mm}
	\centering
	\setlength{\tabcolsep}{3mm}
	\caption{The \emph{Mean Abs 3D Err} comparison of Cano et al. and our results.}\label{tab:Cano}
	\begin{tabular}{cccc}
		\toprule
		Subject            & Video Sequence   & Cano \textsl{et al.}   & Ours  \\
		\midrule
		\multirow{2}*{S1}  &\emph{Gesture}    & 32.3                   & 46.5  \\
				           &\emph{Walking}    & 83.0                   & 53.0  \\
		\multirow{2}*{S2}  &\emph{Gesture}    & 78.4                   & 62.6  \\
                           &\emph{Walking}    & 69.6                   & 59.8  \\			
		\multirow{2}*{S3}  &\emph{Gesture}    & 46.8                   & 61.5  \\
                           &\emph{Walking}    & 111.8                  & 60.3  \\	
		\midrule
        \emph{mean}        &                  & 70.3                   & 57.2  \\
		\bottomrule         
	\end{tabular}
\end{table}

Table \ref{tab:Cano} shows the performance on the comparison results of Cano \textsl{et al.} \cite{Cano2014Parallelization} and our method. 
The result demonstrates that our method can achieve the performance as good with the method of Cano \textsl{et al.}, and our algorithm is more robust for activities involving rapid movements such as \emph{Walking} sequence in S3.

\begin{figure*}[htb]
\centering
	\includegraphics[width=0.92\linewidth]{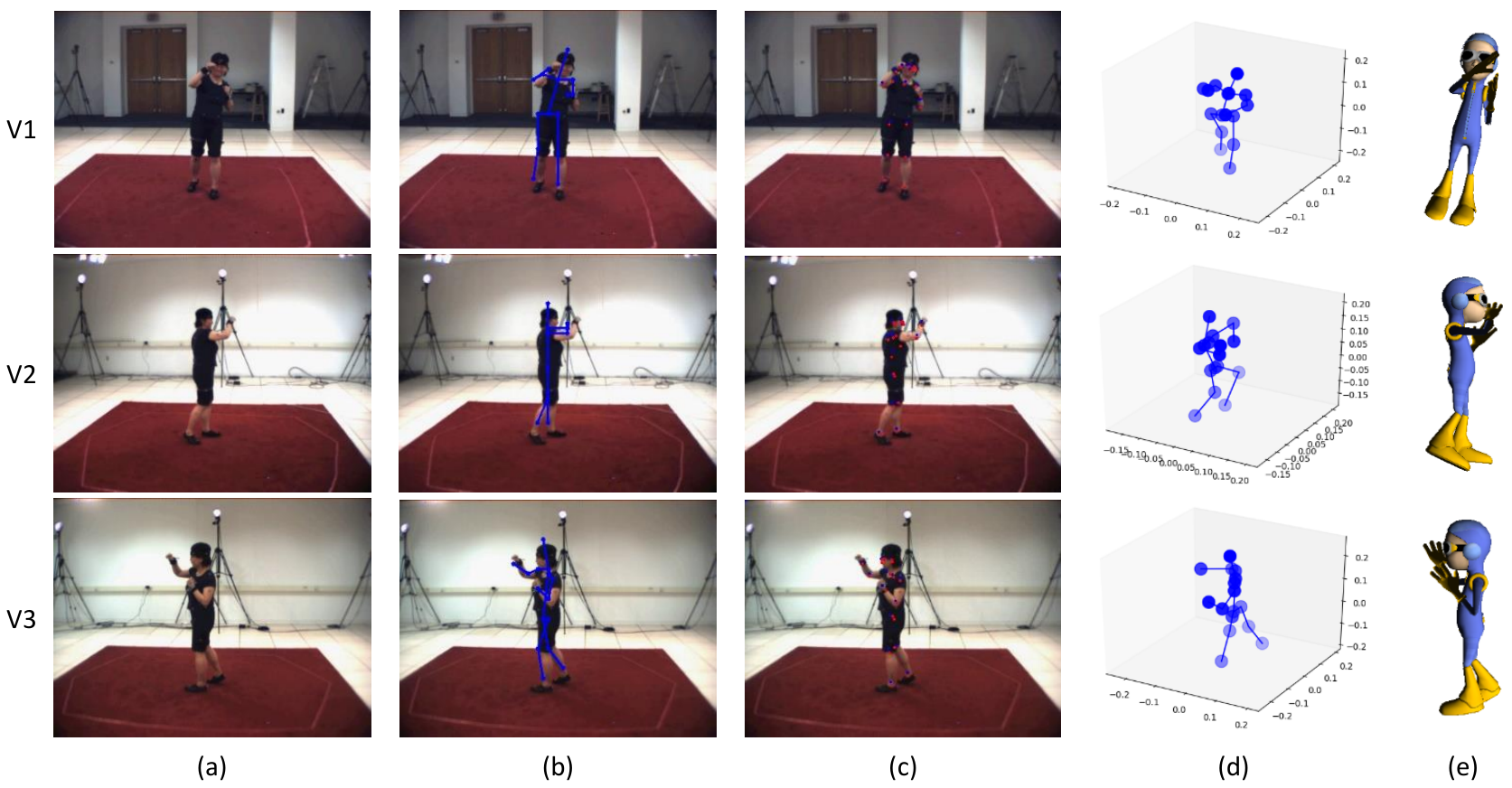}
	\caption{\revised{
		The qualitative comparison of Zhou \textsl{et al.} \cite{Zhou_2017_ICCV} and our method for a specific frame in a video sequence. 
		The rows ``V1'', ``V2'' and ``V3'' are three different views of the same target in the frame.
		The first to fifth columns are: (a) the original images, (b) 2D projected joints (in blue color) of 3D voxel estimated by \cite{Zhou_2017_ICCV}, (c) 2D projected joints of 3D voxel estimated by our method (in red color) and 2D joints detected by deep CNN (in blue color), (d) 3D joints produced by \cite{Zhou_2017_ICCV}, and (e) visualization results generated by our method.
	}}\label{fig:analysis 1}
\end{figure*}

\revised{
Fig. \ref{fig:analysis 1} shows the comparison of Zhou \textsl{et al.} \cite{Zhou_2017_ICCV} and our method for a specific frame of a target in the video sequence from different views.
As the method proposed by Zhou \textsl{et al.} predicts the human 3D joints by a single image, the produced results have a strong correlation with the viewport. 
From the results listed in Fig.\ref{fig:analysis 1}(b) and Fig.\ref{fig:analysis 1}(d) which is produced by the method of \cite{Zhou_2017_ICCV}, it can be found that the estimation results of 3D joints are sensitive to views with errors occurring. It is obviously that there is a big difference between the results of different views, even if the different views are from the same pose of the same target in the video.
Accordingly, Fig.\ref{fig:analysis 1}(c) and (e) indicate that the results generated by our method are overall more robust and accurate. 
}

\begin{figure*}[htb]
	\centering
	\includegraphics[width=0.95\linewidth]{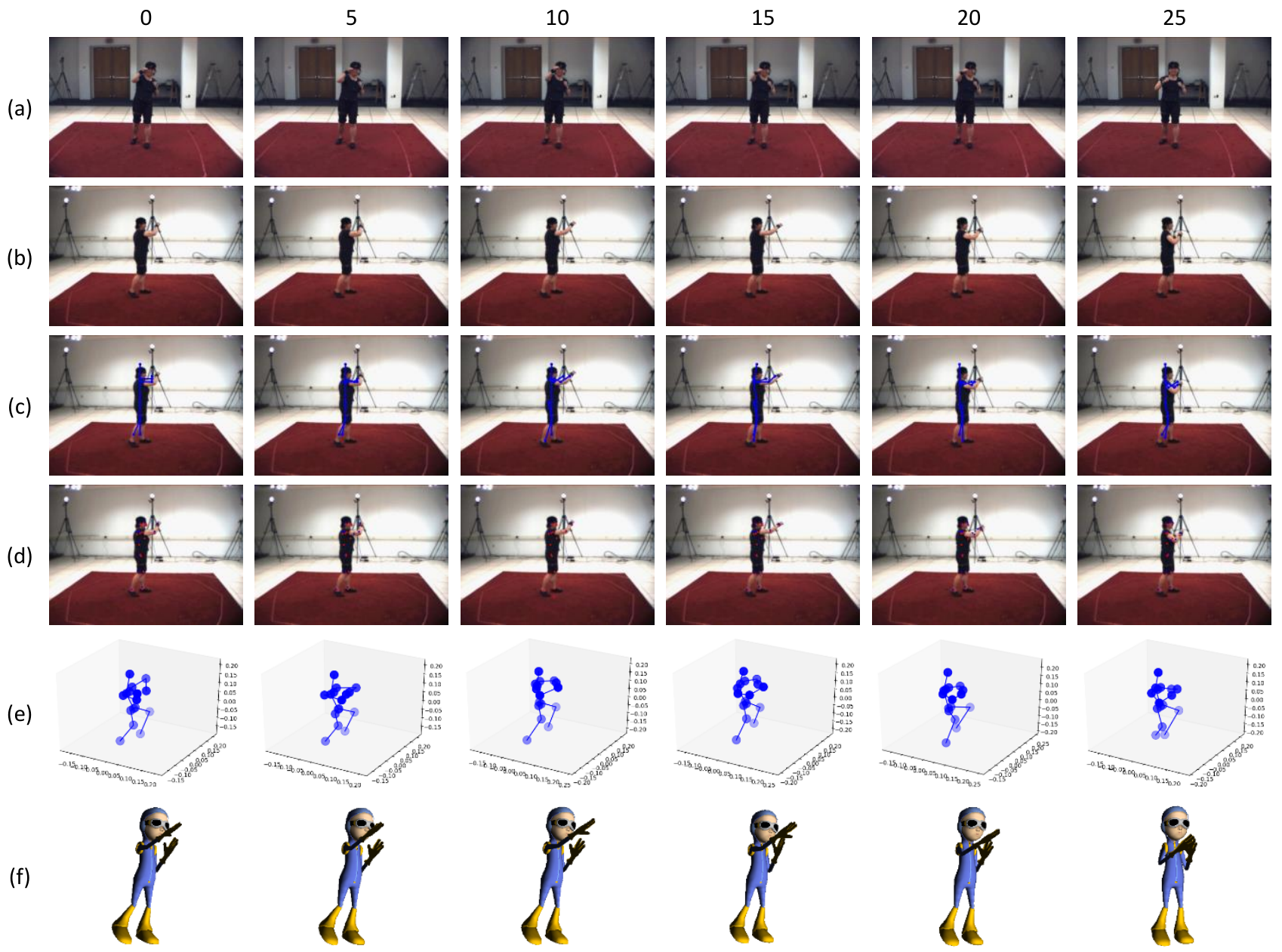}
	\caption{\revised{
			The qualitative comparison II of Zhou \textsl{et al.} \cite{Zhou_2017_ICCV} and our method for continuous video frames. 
			Only some key frames are listed with their indices on the top of the images. The rows from top to bottom are:
			(a) the original images (from front view), (b) the original images (from side view), (c) 2D projected joints (in blue color) of 3D voxel estimated by \cite{Zhou_2017_ICCV}, (d) 2D projected joints of 3D voxel estimated by our method (in red color) and 2D joints detected by deep CNN (in blue color), (e) 3D joints produced by \cite{Zhou_2017_ICCV}, and (f) visualization results generated by our method.
	}}\label{fig:analysis 2}
\end{figure*}

\revised{	
Fig. \ref{fig:analysis 2} shows the comparison of Zhou \textsl{et al.} \cite{Zhou_2017_ICCV} and our method for the continuous video frames of a human action. It can be seen from row (c) and row (e) of Fig. \ref{fig:analysis 2} that the results generated by Zhou \textsl{et al.} drift between frames randomly, which is not acceptable for generating smooth animation, while our method produces more accurate and stable outcomes as shown in row (d) and row (f) of Fig. \ref{fig:analysis 2}.}

\subsection{Experimental analysis} \label{sec:analysis}

\revised{
In this part, we present more analysis about our system itself based on some additional experiments.
}
\subsubsection{\revised{Accuracy analysis of our system}}

\begin{figure*}[htb]
	\centering
	\includegraphics[width=0.98\linewidth]{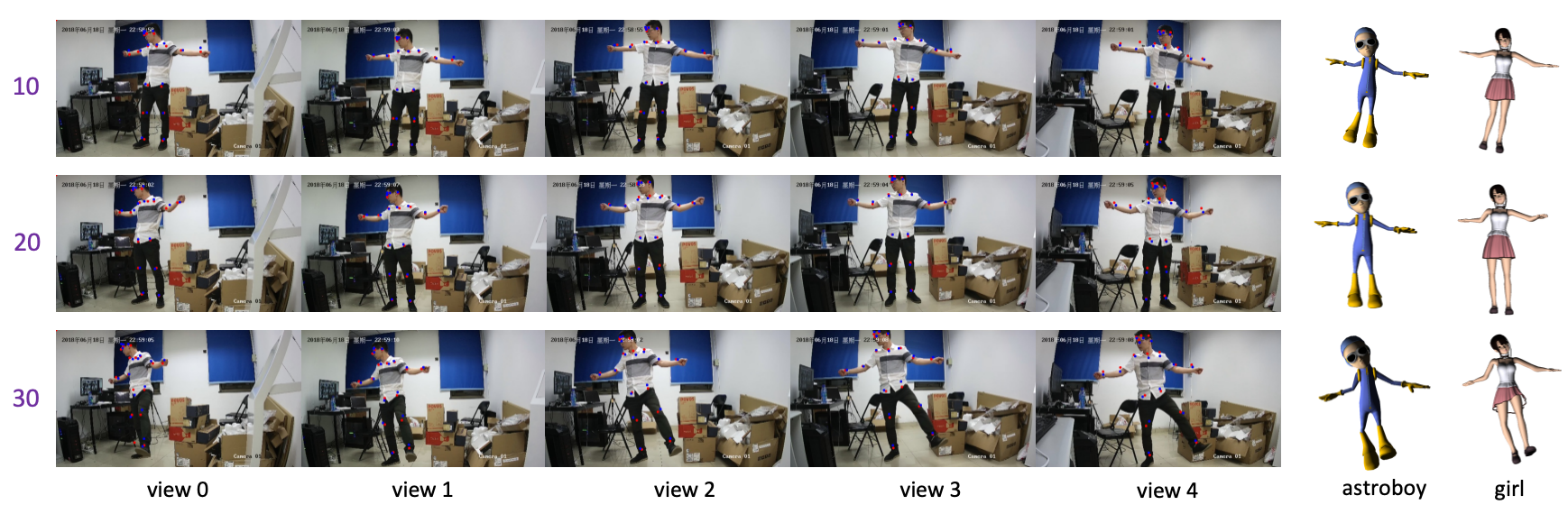}
	\caption{
		Comparison of projection joints of 3D voxel and 2D CNN detection joints.
		10, 20, 30 are the frame of sequence.
		It displays different views from view 0 to view 4.
		Red points are the 2D detection joints by Deep CNN, and blue points are the 2D projection joints of 3D voxel, which is calculated by our method.
		``astroboy'' and ``girl'' are the cartoon characters.
	}
	\label{exp_compare}
\end{figure*}

\begin{table}[th]
	\setlength{\abovecaptionskip}{1mm}
	\centering
	\setlength{\tabcolsep}{1.5mm}
	\caption{The \emph{Avg 2D Err} of the 2D reprojection points from the 3D voxel and 2D joints were detected by the CNN in Fig. \ref{exp_compare}.}\label{Avg 2D Err}
	\begin{tabular}{ccccccc}
		\toprule
		Frame       & View 0 & View 1 & View 2 & View 3 & View 4 & Overall \\
		\midrule
		10	        & 19.7   & 11.8   & 13.8   & 13.2   & 24.3   & 16.5 \\
		20          & 17.9   & 12.6   & 12.4   & 9.7    & 23.3   & 15.1 \\
		30          & 28.1   & 18.0   & 18.6   & 16.2   & 23.7   & 20.9 \\
		\midrule
		\emph{mean} & 21.9   & 14.1   & 14.9   & 13.0   & 23.7   & 17.5 \\
		\bottomrule
	\end{tabular}
\end{table}

\begin{figure}[th]
	\centering
	\includegraphics[width=0.98\linewidth]{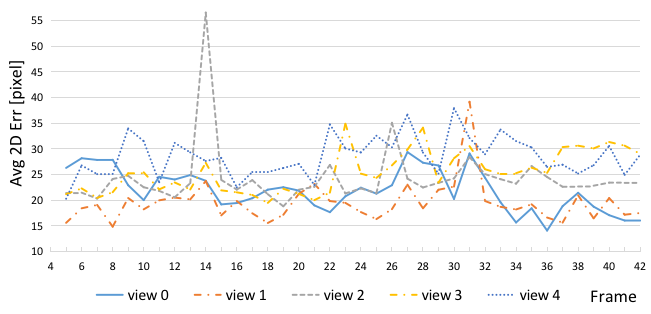}
	\caption{
		The \emph{Avg 2D Err} in multi-views.
		In general, the average 2D error of each view is kept within a small ranges.
		The very high intensity is the noise, when 3D points or 2D CNN points are not detected.
		Frames 1-4 represent the case that no 3D points are detected. 
	}
	\label{2d err}
\end{figure}

In order to evaluate \revised{the accuracy of} our method, we define the average 2D error (\emph{Avg 2D Err}) directly in the image in pixels (pix).
The \emph{Avg 2D Err} between the estimated pose $ x $ and the ground truth pose $ \hat{x} $ is expressed as the average Euclidean distance \cite{Sigal2010HumanEva} between individual markers: 
\begin{equation}\label{formula_2d}
	d^2(x, \hat{x}) = \frac{1}{n}\sum_{i=1}^n||x_i - \hat{x}_i||,
\end{equation}
where $ n $ is the number of joints, $ i = 1 \hdots n$, $ x $ is the 2D joint directly obtained from the 2D CNN detector, and $ \hat{x} $ is the 2D reprojected joint of the calculated 3D joint.
Fig. \ref{exp_compare} shows the comparison of the reprojected points between the calculated 3D joint points and the results directly obtained by the 2D CNN detector. 
Table \ref{Avg 2D Err} presents the results of calculating the \emph{Avg 2D Err} to evaluate the effectiveness of the reconstruction joint in 3D space.
The value (pixel distance) of each view is the average error of all 2D joints. 
Here, the image resolution is $ 1920 \times 1080 $.
The ratio of \emph{Avg 2D Err} and image resolution is under 2\%.
The accuracy of the cameras' intrinsic and extrinsic parameters was not good, and this led to the \emph{Avg 2D Err} ranges by view.
From the comparison of one view, such as frame 10, view 0 in Fig. \ref{exp_compare}, it can be seen that the proposed 3D voxel joint estimation method can precisely reconstruct the key joints of the body.
Additionally, the bone transformation calculation can obtain the correct bone pose in 3D space, and the results can be used to animate multiple 3D characters.
\revised{Fig. \ref{2d err} shows the multi-views \emph{Avg 2D Err} value of consecutive frames in our example video sequence as shown in Fig. \ref{exp_compare}. It shows that our 3D joint estimation method is stable and reliable.}



\subsubsection{\revised{Time consumption analysis of the whole pipeline}}

\begin{table*}[th]
	\setlength{\abovecaptionskip}{1mm}
	\centering
	\setlength{\tabcolsep}{4mm}
	\caption{\revised{The time consumption of the 4 main phases in the whole pipeline. 
	}}\label{tab:pipeline time consumption}
	\begin{tabular}{lcccc}
		\toprule
		\multirow{2}*{Phase} & Image Collection      & 2D Joint Detection        & 3D Joint Estimation        & Character Animation  \\
		                     & (5 views)             & (once for a merged image) & (18 joints, non-parallel)  & (rigging \& skinning) \\
		\midrule
		Time (ms)            & $50$ (\emph{average}) & $120$ (\emph{average})    & $100$ (\emph{average})     & $ < 5 $    \\
		\bottomrule
	\end{tabular}
\end{table*}

\revised{
Table \ref{tab:pipeline time consumption} shows the time consumption of our pipeline in millisecond (ms) by 4 main phases.
The image collection phase costs a lot for image preprocessing in a high frame rate (about $30$fps). 
The second phase (2D joint detection) takes about $120$ms, which is the maximum in all the phases, because the deep CNN is very time-consuming due to many large computational processes. Nevertheless, the 2D joint detection using CNN is still much faster than the method based on feature correspondence.
Additionally, another time-consuming phase is 3D joint estimation, in which 18 joints are calculated one after another and the average processing time of each joint is about $5.6$ms.
The final phase is for generating character animation, including rigging and skinning, which usually takes less than $5$ms.
For now, the frame rate of our system can achieve $5$fps by a preliminary implementation without any optimization in the experimental environment. 
In the future, the frame rate could be greatly improved by following strategies: optimizing the network structure of the deep CNN for 2D joints detection and adopting parallel computing techniques for 3D joints estimation.
}

\subsubsection{\revised{The effect of the resolution of input images}}


\begin{table}[th]
	\setlength{\abovecaptionskip}{1mm}
	\centering
	\setlength{\tabcolsep}{3mm}
	\caption{\revised{The experimental results about the effect of down-sampling and merging input images.}}\label{tab:effect of down-sample}
	\begin{tabular}{lcc}
       	\toprule
       	Merging \& Down-sampling    & No                 &  Yes               \\
       	\midrule
       	Resolution (pix)            & $640 \times 480$   & $427 \times 480$ \\
  		Time (ms)                   & 633                &  \textbf{181}               \\
  		\emph{Mean Abs 3D Err} (mm) & 38.6               &  52.1              \\
       \bottomrule
   \end{tabular}
\end{table}

\revised{
In the process of 2D joint detection described in Section \ref{sec:stage_2}, the input images captured by cameras are down-sampled and merged into a single image.
In order to analyze the effect on the performance of 3D joint estimation introduced by down-sampling and merging input images, an additional experiment was carried out which processed 5 original input images separately, and the time consumption and accuracy of the results were compared with those of the proposed method. In the experiment, $\delta$ and $\sigma$ were set to $[20, 20, 20]$ and $4$ respectively, and \emph{Box} sequence of S1 in the HumanEva-I dataset was chosen as the test data. From the results listed in Table \ref{tab:effect of down-sample}, it can be found that the input image resolution is responsible for the 3D joint estimation accuracy. Indeed, down-sampling operation will lead to a little increase in the 3D joint estimation error. However, merging input images into one big image makes the time consumption of the pipeline reduced dramatically. Generally speaking, it's worth sacrificing a little accuracy to improve efficiency greatly.}

\subsubsection{\revised{More analysis of 3D joint estimation}}


\begin{table}[th]
	\setlength{\abovecaptionskip}{1mm}
	\centering
	\setlength{\tabcolsep}{2.1mm}
	\caption{\revised{The time consumption and \emph{Mean Abs 3D Err} results of \emph{Box} squence of S1 in the HumanEva-I dataset.}}\label{tab:time consumption}
	\begin{tabular}{ccc}
		\toprule
		  $\delta$ (mm)    & Time (ms) & \emph{Mean Abs 3D Err} (mm) \\
		\midrule
		  $[1, 1, 1]$      & 6249    & 68.3 \\
		  $[3, 3, 3]$      & 985     & 54.2 \\
		  $[5, 5, 5]$      & 218     & 52.7 \\
		  $[10, 10, 10]$   & 74      & 52.3 \\
		  $[15, 15, 15]$   & 69      & 52.3 \\
		  $[20, 20, 20]$   & 34      & 52.1 \\
		  $[25, 25, 25]$   & 33      & 52.1 \\
		  $[30, 30, 30]$   & 33     & 52.1\\
		\bottomrule
	\end{tabular}
\end{table}

\revised{
The time consumption and result quality (evaluated by \emph{Mean Abs 3D Err}) of the 3D joint estimation phase were analyzed by a further experiment with different $\delta$ settings, in which the merged image resolution was set to $1280\times 960$ and $\sigma$ was set to $4$.
From the results listed in Table \ref{tab:time consumption}, it can be found that the sample cube size $\delta$ plays an important role in terms of time consumption. When $\delta$ decreases, the time consumption increases dramatically.
However, \emph{Mean Abs 3D Err} is not sensitive to the change of $\delta$. When $\delta$ alters, \emph{Mean Abs 3D Err} doesn't change much. Therefore, it is not difficult to find a trade-off between computing precision and speed with appropriate parameters.
}
 
\section{Conclusions}\label{sec:conclusions}
This paper proposes a new system for markerless human body motion capturing and animated character rigging using multi-view cameras. According to the experimental results, our system can produce accurate and robust 3D human body joints from multi-view camera images, which can then be used to rig 3D characters for animation. This system may be used in fields such as animation production, video game production, and VR game interaction. Thus, the production costs can be reduced and the human-machine interaction can be simplified considerably.
However, the development of the proposed system is in the prototype stage, and there still exist issues that require further investigation, such as the stability of the calculated 3D joints, number and layout of the cameras, and efficiency of the entire system.
In future work, we will focus on improving the reconstruction of the 3D joints and the capturing of body motion.

\bibliographystyle{IEEEtran} 
\bibliography{reference}

\begin{thebibliography}{10}
\providecommand{\url}[1]{#1}
\csname url@samestyle\endcsname
\providecommand{\newblock}{\relax}
\providecommand{\bibinfo}[2]{#2}
\providecommand{\BIBentrySTDinterwordspacing}{\spaceskip=0pt\relax}
\providecommand{\BIBentryALTinterwordstretchfactor}{4}
\providecommand{\BIBentryALTinterwordspacing}{\spaceskip=\fontdimen2\font plus
\BIBentryALTinterwordstretchfactor\fontdimen3\font minus
  \fontdimen4\font\relax}
\providecommand{\BIBforeignlanguage}[2]{{%
\expandafter\ifx\csname l@#1\endcsname\relax
\typeout{** WARNING: IEEEtran.bst: No hyphenation pattern has been}%
\typeout{** loaded for the language `#1'. Using the pattern for}%
\typeout{** the default language instead.}%
\else
\language=\csname l@#1\endcsname
\fi
#2}}
\providecommand{\BIBdecl}{\relax}
\BIBdecl

\bibitem{Moeslund2006A}
T.~B. Moeslund, A.~Hilton, and V.~Kr{\"u}ger, ``A survey of advances in
  vision-based human motion capture and analysis,'' \emph{Computer Vision \&
  Image Understanding}, vol. 104, no.~2, pp. 90--126, 2006.

\bibitem{Magnor2015Digital}
M.~A. Magnor, O.~Grau, O.~Sorkinehornung, and C.~Theobalt, ``Digital
  representations of the real world: How to capture, model, and render visual
  reality,'' 2015.

\bibitem{Bezodis2008Lower}
I.~N. Bezodis, D.~G. Kerwin, and A.~I. Salo, ``Lower-limb mechanics during the
  support phase of maximum-velocity sprint running,'' \emph{Med Sci Sports
  Exerc}, vol.~40, no.~4, pp. 707--715, 2008.

\bibitem{Churchill2015The}
S.~M. Churchill, A.~I.~T. Salo, and G.~Trewartha, ``The effect of the bend on
  technique and performance during maximal effort sprinting,'' \emph{Sports
  Biomechanics}, vol.~14, no.~1, pp. 106--121, 2015.

\bibitem{Hiley2012Achieving}
M.~J. Hiley and M.~R. Yeadon, ``Achieving consistent performance in a complex
  whole body movement: the tkatchev on high bar,'' \emph{Human Movement
  Science}, vol.~31, no.~4, pp. 834--843, 2012.

\bibitem{Neil2007Contributions}
N.~Bezodis, G.~Trewartha, C.~Wilson, and G.~Irwin, ``Contributions of the
  non-kicking-side arm to rugby place-kicking technique,'' \emph{Sports
  Biomechanics}, vol.~6, no.~2, pp. 171--186, 2007.

\bibitem{Baran2007Automatic}
I.~Baran, ``Automatic rigging and animation of 3d characters,'' \emph{ACM
  Transactions on Graphics (TOG)}, vol.~26, no.~3, p.~72, 2007.

\bibitem{Raskar2007Prakash}
R.~Raskar, H.~Nii, B.~Dedecker, Y.~Hashimoto, J.~Summet, D.~Moore, Y.~Zhao,
  J.~Westhues, P.~Dietz, and J.~Barnwell, ``Prakash:lighting aware motion
  capture using photosensing markers and multiplexed illuminators,'' \emph{Acm
  Transactions on Graphics}, vol.~26, no.~3, p.~36, 2007.

\bibitem{Shiratori2011Motion}
T.~Shiratori, L.~Sigal, L.~Sigal, Y.~Sheikh, and J.~K. Hodgins, ``Motion
  capture from body-mounted cameras,'' in \emph{ACM SIGGRAPH}, 2011, p.~31.

\bibitem{Della2005Human}
C.~U. Della, A.~Leardini, L.~Chiari, and A.~Cappozzo, ``Human movement analysis
  using stereophotogrammetry. part 3: Soft tissue artifact assessment and
  compensation,'' \emph{Gait \& Posture}, vol.~21, no.~2, pp. 221--225, 2005.

\bibitem{Richards1999The}
J.~G. Richards, ``The measurement of human motion: A comparison of commercially
  available systems,'' \emph{Human Movement Science}, vol.~18, no.~5, pp.
  589--602, 1999.

\bibitem{Colyer2018A}
S.~L. Colyer, M.~Evans, D.~P. Cosker, and A.~I.~T. Salo, ``A review of the
  evolution of vision-based motion analysis and the integration of advanced
  computer vision methods towards developing a markerless system,''
  \emph{Sports Medicine - Open}, vol.~4, no.~1, p.~24, 2018.

\bibitem{Shotton2011Real}
J.~Shotton, A.~Fitzgibbon, M.~Cook, and T.~Sharp, ``Real-time human pose
  recognition in parts from single depth images,'' in \emph{IEEE Conference on
  Computer Vision and Pattern Recognition}, 2011, pp. 1297--1304.

\bibitem{Ye2013A}
M.~Ye, Q.~Zhang, L.~Wang, J.~Zhu, R.~Yang, and J.~Gall, \emph{A Survey on Human
  Motion Analysis from Depth Data}.\hskip 1em plus 0.5em minus 0.4em\relax
  Springer Berlin Heidelberg, 2013.

\bibitem{Ganapathi2010Real}
V.~Ganapathi, C.~Plagemann, D.~Koller, and S.~Thrun, ``Real time motion capture
  using a single time-of-flight camera,'' in \emph{Computer Vision and Pattern
  Recognition}, 2010, pp. 755--762.

\bibitem{Shotton2013Real}
J.~Shotton, T.~Sharp, A.~Kipman, A.~Fitzgibbon, M.~Finocchio, A.~Blake,
  M.~Cook, and R.~Moore, \emph{Real-time human pose recognition in parts from
  single depth images}.\hskip 1em plus 0.5em minus 0.4em\relax Springer Berlin
  Heidelberg, 2013.

\bibitem{Cao2016Realtime}
Z.~Cao, T.~Simon, S.~E. Wei, and Y.~Sheikh, ``Realtime multi-person 2d pose
  estimation using part affinity fields,'' pp. 1302--1310, 2016.

\bibitem{Gall2009Motion}
J.~Gall, C.~Stoll, E.~D. Aguiar, C.~Theobalt, B.~Rosenhahn, and H.~P. Seidel,
  ``Motion capture using joint skeleton tracking and surface estimation,'' in
  \emph{Computer Vision and Pattern Recognition, 2009. CVPR 2009. IEEE
  Conference on}, 2009, pp. 1746--1753.

\bibitem{Bregler2004Twist}
C.~Bregler, J.~Malik, and K.~Pullen, ``Twist based acquisition and tracking of
  animal and human kinematics,'' \emph{International Journal of Computer
  Vision}, vol.~56, no.~3, pp. 179--194, 2004.

\bibitem{Duetscher2000Articulated}
J.~Duetscher, A.~Blake, and I.~Reid, ``Articulated body motion capture by
  annealed particle filtering,'' in \emph{Computer Vision and Pattern
  Recognition, 2000. Proceedings. IEEE Conference on}, 2000, pp. 126--133
  vol.2.

\bibitem{Agarwal2005Recovering}
A.~Agarwal and B.~Triggs, ``Recovering 3d human pose from monocular images,''
  \emph{IEEE Transactions on Pattern Analysis \& Machine Intelligence},
  vol.~28, no.~1, pp. 44--58, 2005.

\bibitem{Hong2016Multi}
C.~Hong, J.~Yu, Y.~Xie, and X.~Chen, ``Multi-view deep learning for image-based
  pose recovery,'' in \emph{IEEE International Conference on Communication
  Technology}, 2016, pp. 897--902.

\bibitem{Chen20113D}
C.~Chen, Y.~Yang, F.~Nie, and J.~M. Odobez, ``3d human pose recovery from image
  by efficient visual feature selection,'' \emph{Computer Vision \& Image
  Understanding}, vol. 115, no.~3, pp. 290--299, 2011.

\bibitem{Babagholami2014A}
B.~Babagholami-Mohamadabadi, A.~Jourabloo, A.~Zarghami, and S.~Kasaei, ``A
  bayesian framework for sparse representation-based 3-d human pose
  estimation,'' \emph{IEEE Signal Processing Letters}, vol.~21, no.~3, pp.
  297--300, 2014.

\bibitem{Theobalt2017VNect}
Christian and Theobalt, ``Vnect: real-time 3d human pose estimation with a
  single rgb camera,'' \emph{Acm Transactions on Graphics}, vol.~36, no.~4,
  p.~44, 2017.

\bibitem{Rhodin2016General}
H.~Rhodin, N.~Robertini, C.~Dan, C.~Richardt, H.~P. Seidel, and C.~Theobalt,
  \emph{General Automatic Human Shape and Motion Capture Using Volumetric
  Contour Cues}.\hskip 1em plus 0.5em minus 0.4em\relax Springer International
  Publishing, 2016.

\bibitem{Corazza2010Markerless}
S.~Corazza, L.~M{\"u}ndermann, E.~Gambaretto, G.~Ferrigno, and T.~P.
  Andriacchi, ``Markerless motion capture through visual hull, articulated icp
  and subject specific model generation,'' \emph{International Journal of
  Computer Vision}, vol.~87, no. 1-2, p. 156, 2010.

\bibitem{Corazza2006A}
S.~Corazza, A.~M. Mundermann~LChaudhari, T.~Demattio, C.~Cobelli, and T.~P.
  Andriacchi, ``A markerless motion capture system to study musculoskeletal
  biomechanics: visual hull and simulated annealing approach.'' \emph{Annals of
  Biomedical Engineering}, vol.~34, no.~6, pp. 1019--1029, 2006.

\bibitem{Stoll2011Fast}
C.~Stoll, N.~Hasler, J.~Gall, H.~P. Seidel, and C.~Theobalt, ``Fast articulated
  motion tracking using a sums of gaussians body model,'' in
  \emph{International Conference on Computer Vision}, 2011, pp. 951--958.

\bibitem{De2008Performance}
E.~De~Aguiar, C.~Stoll, C.~Theobalt, N.~Ahmed, H.~P. Seidel, and S.~Thrun,
  ``Performance capture from sparse multi-view video,'' \emph{Acm Transactions
  on Graphics}, vol.~27, no.~3, pp. 1--10, 2008.

\bibitem{Guo2015Robust}
K.~Guo, F.~Xu, Y.~Wang, and Y.~Liu, ``Robust non-rigid motion tracking and
  surface reconstruction using l0 regularization,'' in \emph{IEEE International
  Conference on Computer Vision}, 2015, pp. 3083--3091.

\bibitem{Vlasic2008Articulated}
D.~Vlasic, I.~Baran, and W.~Matusik, ``Articulated mesh animation from
  multi-view silhouettes,'' \emph{Acm Transactions on Graphics}, vol.~27,
  no.~3, pp. 1--9, 2008.

\bibitem{Hasler2009Markerless}
N.~Hasler, B.~Rosenhahn, T.~Thormahlen, M.~Wand, J.~Gall, and H.~P. Seidel,
  ``Markerless motion capture with unsynchronized moving cameras,'' in
  \emph{Computer Vision and Pattern Recognition, 2009. CVPR 2009. IEEE
  Conference on}, 2009, pp. 224--231.

\bibitem{Ye2012Performance}
G.~Ye, Y.~Liu, N.~Hasler, X.~Ji, Q.~Dai, and C.~Theobalt, ``Performance capture
  of interacting characters with handheld kinects,'' in \emph{European
  Conference on Computer Vision}, 2012, pp. 828--841.

\bibitem{Liu2011Markerless}
Y.~Liu, C.~Stoll, J.~Gall, H.~P. Seidel, and C.~Theobalt, ``Markerless motion
  capture of interacting characters using multi-view image segmentation,'' in
  \emph{Computer Vision and Pattern Recognition}, 2011, pp. 1249--1256.

\bibitem{Liu2013Markerless}
Y.~Liu, J.~Gall, C.~Stoll, Q.~Dai, H.~P. Seidel, and C.~Theobalt, ``Markerless
  motion capture of multiple characters using multiview image segmentation.''
  \emph{IEEE Transactions on Pattern Analysis \& Machine Intelligence},
  vol.~35, no.~11, pp. 2720--2735, 2013.

\bibitem{Flam2009OpenMoCap}
D.~L. Flam, D.~P.~D. Queiroz, A.~D.~A. Ara{\'u}jo, and J.~V.~B. Gomide,
  \emph{OpenMoCap: An Open Source Software for Optical Motion Capture}.\hskip
  1em plus 0.5em minus 0.4em\relax IEEE Computer Society, 2009.

\bibitem{Xu2016FlyCap}
L.~Xu, Y.~Liu, W.~Cheng, K.~Guo, G.~Zhou, Q.~Dai, and L.~Fang, ``Flycap:
  Markerless motion capture using multiple autonomous flying cameras,''
  \emph{IEEE Transactions on Visualization \& Computer Graphics}, vol.~PP,
  no.~99, pp. 1--1, 2016.

\bibitem{Pavlakos2017Harvesting}
G.~Pavlakos, X.~Zhou, K.~G. Derpanis, and K.~Daniilidis, ``Harvesting multiple
  views for marker-less 3d human pose annotations,'' in \emph{IEEE Conference
  on Computer Vision and Pattern Recognition}, 2017, pp. 1253--1262.

\bibitem{Triggs1999Bundle}
B.~Triggs, P.~F. Mclauchlan, R.~I. Hartley, and A.~W. Fitzgibbon, \emph{Bundle
  Adjustment — A Modern Synthesis}, 1999.

\bibitem{Cano2014Parallelization}
A.~Cano, E.~Yeguas-Bolivar, R.~Muñoz-Salinas, R.~Medina-Carnicer, and
  S.~Ventura, ``Parallelization strategies for markerless human motion
  capture,'' \emph{Journal of Real-Time Image Processing}, vol.~14, no.~2, pp.
  1--15, 2014.

\bibitem{Zhou_2017_ICCV}
X.~Zhou, Q.~Huang, X.~Sun, X.~Xue, and Y.~Wei, ``Towards 3d human pose
  estimation in the wild: A weakly-supervised approach,'' in \emph{The IEEE
  International Conference on Computer Vision (ICCV)}, Oct 2017.

\bibitem{Sigal2010HumanEva}
L.~Sigal, A.~O. Balan, and M.~J. Black, ``Humaneva: Synchronized video and
  motion capture dataset and baseline algorithm for evaluation of articulated
  human motion,'' \emph{International Journal of Computer Vision}, vol.~87, no.
  1-2, p.~4, 2010.

\end{thebibliography}

\begin{IEEEbiography}[{\includegraphics[width=1in,height=1.25in,clip,keepaspectratio]{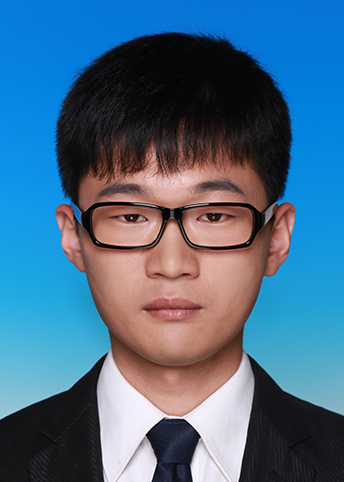}}]{Jinbao Wang}
received the B.S. degree from Hebei University in 2013 and the M.S. degree from Beijing Union University in 2016. He is now a Ph.D. student of University of Chinese Academy of Sciences. His research interests include digital image processing and computer vision.
\end{IEEEbiography}

\begin{IEEEbiography}[{\includegraphics[width=1in,height=1.25in,clip,keepaspectratio]{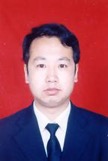}}]{Ke Lu}
was born in Ningxia on March 13th, 1971. He received master degree and Ph.D. degree from the Department of Mathematics and Department of Computer Science at Northwest University in July 1998 and July 2003, respectively. He worked as a postdoctoral fellow in the Institute of Automation Chinese Academy of Sciences from July 2003 to April 2005. Currently he is a professor of the University of the Chinese Academy of Sciences. His current research areas focus on computer vision, 3D image reconstruction and computer graphics.
\end{IEEEbiography}

\begin{IEEEbiography}[{\includegraphics[width=1in,height=1.25in,clip,keepaspectratio]{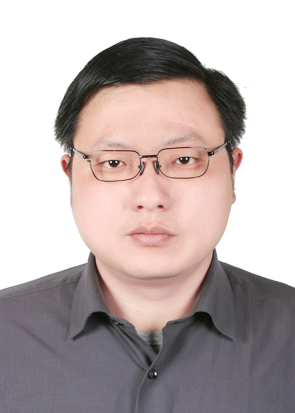}}]{Jian Xue}
was born in Jiangsu, China, in 1979. He received the Ph.D. degree in computer applied technology from the Institute of Automation, Chinese Academy of Sciences, Beijing, China, in July 2007. He worked as an academic visitor in the University of Western Australia from July to December, 2012. Currently he is an Associate Professor with the University of Chinese Academy of Sciences, Beijing. His current research interests include image processing, computer graphics and scientific visualization.
\end{IEEEbiography}

\vfill

\end{document}